\documentclass[10pt,twocolumn,letterpaper]{article}
\pdfoutput=1
\usepackage{cvpr}
\usepackage{times}
\usepackage{epsfig}
\usepackage{graphicx}
\usepackage{amsmath}
\usepackage{amssymb}
\usepackage{subfigure}
\usepackage{caption}
\usepackage{amssymb}
\usepackage{pifont}
\graphicspath{{./}{./figure/}}
\DeclareGraphicsExtensions{.pdf,.jpeg,.png,.jpg}

\usepackage{hyperref}
\hypersetup{pagebackref=true,breaklinks=true,letterpaper=true,colorlinks,bookmarks=false}
\cvprfinalcopy 

\ifcvprfinal\fi

\begin{document}

\title{Consistent Optimization for Single-Shot Object Detection}

\author{Tao Kong$^{1\dag}$ ~~ Fuchun Sun$^{1}$ ~~ Huaping Liu$^{1}$ ~~Yuning Jiang$^{2}$ ~~ Jianbo Shi$^{3}$\\
$^{1}$Department of Computer Science and Technology, Tsinghua University, \\
Beijing National Research Center for Information Science and Technology (BNRist)\\
$^{2}$ByteDance AI Lab ~~ $^{3}$University of Pennsylvania \\
\tt\small taokongcn@gmail.com, \{fcsun,hpliu\}@tsinghua.edu.cn, \\
\tt\small jiangyuning@bytedance.com, jshi@seas.upenn.edu
}

\maketitle

\begin{abstract}
   We present consistent optimization for single stage object detection. Previous works of single stage object detectors usually rely on the regular, dense sampled anchors to generate hypothesis for the optimization of the model. Through an examination of the behavior of the detector, we observe that the misalignment between the optimization target and inference configurations has hindered the performance improvement. We propose to bride this gap by consistent optimization, which is an extension of the traditional single stage detector's optimization strategy. Consistent optimization focuses on matching the training hypotheses and the inference quality by utilizing of the refined anchors during training.

   To evaluate its effectiveness, we conduct various design choices based on the state-of-the-art RetinaNet detector. We demonstrate it is the consistent optimization, not the architecture design, that yields the performance boosts. Consistent optimization is nearly cost-free, and achieves stable performance gains independent of the model capacities or input scales. Specifically, utilizing consistent optimization improves  RetinaNet from 39.1 AP to 40.1 AP on COCO dataset without any bells or whistles, which surpasses the accuracy of
all existing state-of-the-art one-stage detectors when adopting ResNet-101 as backbone. The code will be made available.
   \let\thefootnote\relax\footnotetext{$^\dag$Part of the work was done when Tao Kong was a visiting scholar at University of Pennsylvania.}

\end{abstract}

\section{Introduction}

\begin{figure}[t]
\centering
\includegraphics[width=0.9\linewidth]{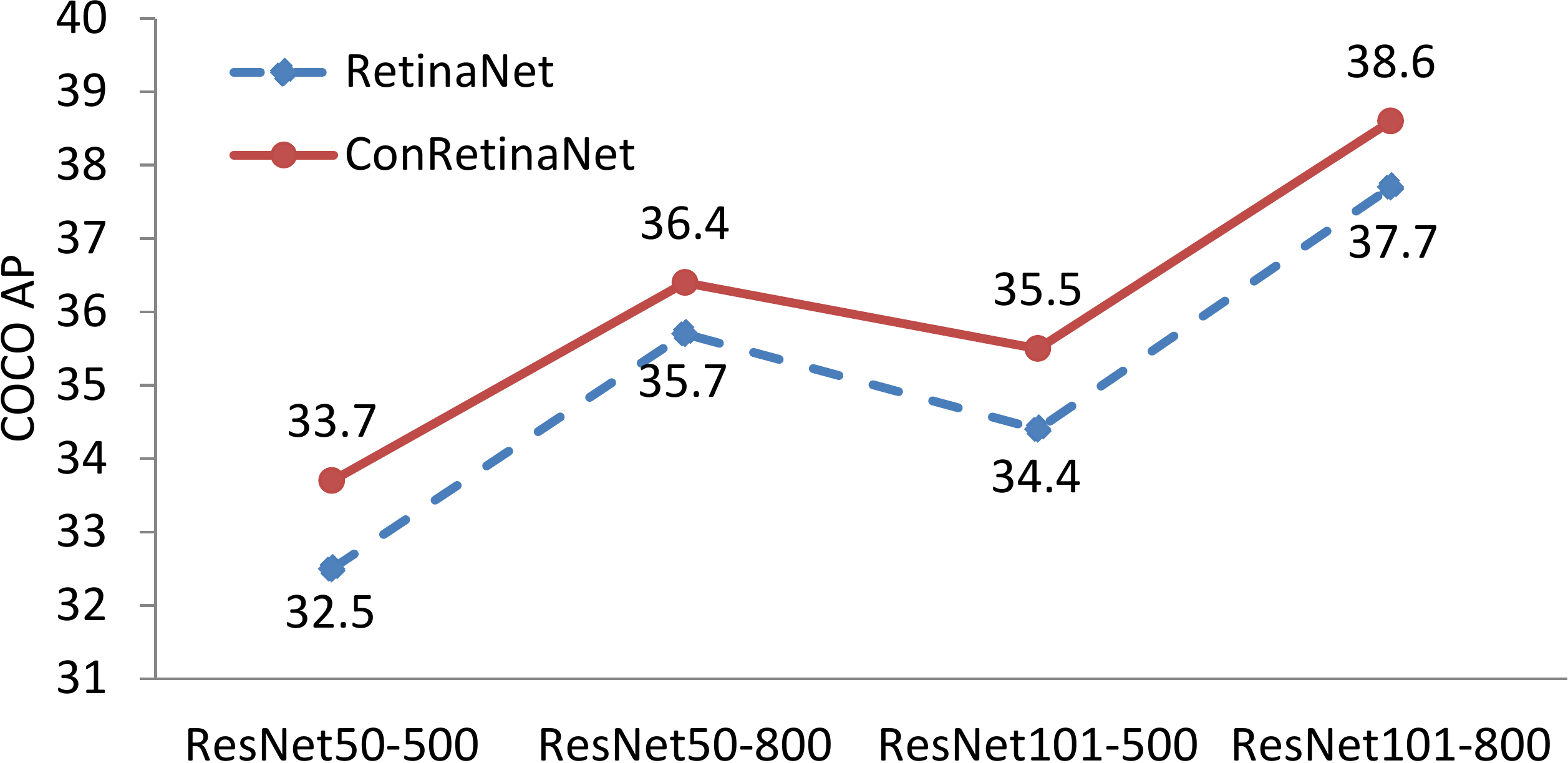}
\caption{RetinaNet \emph{v.s.} ConRetinaNet on different model capacities and input resolutions.  An improved
variant of ConRetinaNet achieves 40.1 AP with ResNet-101 backbone, which is not shown in this figure. Details are given in \textsection\ref{sec:exp}.}
\label{fig:conretinanet}
\end{figure}

We are witnessing the remarkable progress for object detection by exploring the powerful deep leaning technology.
Current state-of-the-art deep learning based object detection frameworks can be generally divided into two major groups:
two-stage, proposal-driven methods \cite{fasterrcnn}\cite{fastrcnn}\cite{fpn} and one-stage, proposal-free methods \cite{ssd}\cite{yolo}. Compared with the two-stage framework, one-stage object detector is more difficult, since it relies on a powerful model head to predict the regular, dense sampled anchors at different locations, scales, and aspect ratios.
The main advantage of one-stage detector is its high computational efficiency. However, its detection accuracy is usually behind that of the two-stage approach, one of the main reasons being due
to the class imbalance problem \cite{retinanet}\cite{ron}.
The most recent work of RetinaNet \cite{retinanet}, is able to match the accuracy of the existing state-of-the-art two-stage detectors, by utilizing Focal Loss to address the \emph{foreground-background class imbalance} challenge.

\begin{figure}[t]
\centering
\subfigure[]{
\includegraphics[height=3cm]{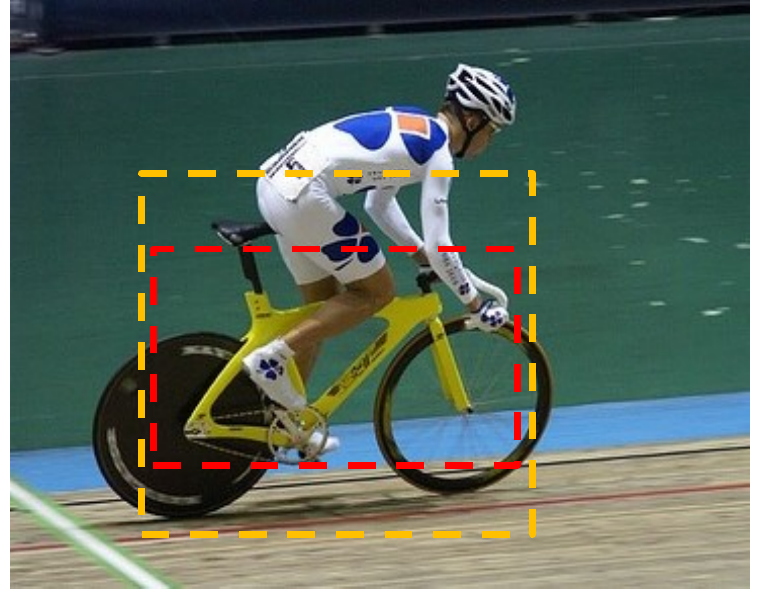}
\label{fig:demo_a}
}
\subfigure[]{
\includegraphics[height=3cm]{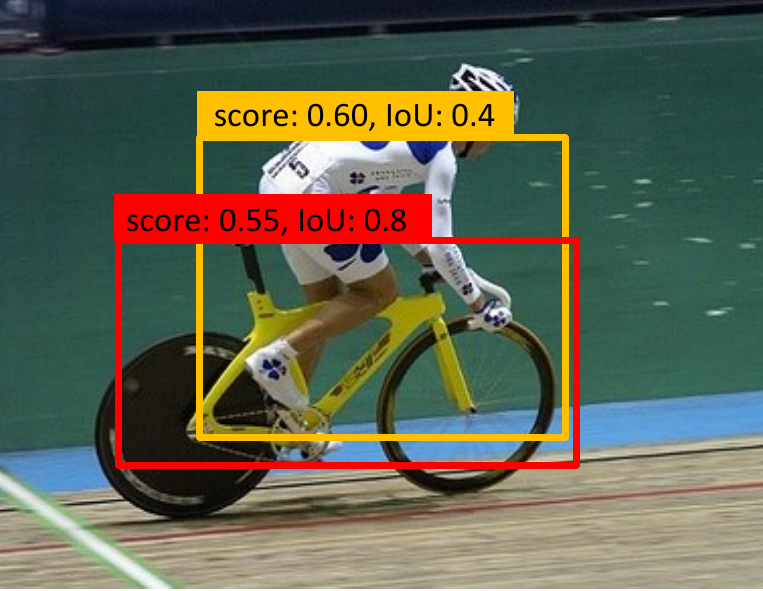}
\label{fig:demo_b}
}
\caption{Demonstrative case of the misalignment between training target and predicted results. (a) There are two anchors matching the groundtruth \emph{bicycle}. (b) The resulting scores and IoUs of the corresponding refined anchors for class \emph{bicycle}. The red bounding box will be suppressed during NMS procedure.}
\label{fig:demo}
\end{figure}

In this work, we observe that in addition to the foreground-background class imbalance challenge, current single shot object detectors also face another challenge: \emph{the misalignment between the training targets and inference configurations}. Given the candidate anchors, the one-stage object detector combines both object \emph{classification} and \emph{localization}. At training phase, the goal of classification sub-networks is to assign the candidate anchor to one of $M$+ 1 classes,  where class 0 contains background and the
remaining the objects to detect. The localization module finds the optimal transformation for the anchor to best fit the groundtruth. At inference, for each anchor, the location is refined by the regression sub-networks. The class probability of the refined anchor is given by the classification sub-networks. The misalignment is: \emph{the training target of the classification is to classify the default, regular anchor, while the predicted probability is assigned to the corresponding regressed anchor which is generated by the localization branch}. 

\begin{figure}[!h]
\centering
\includegraphics[height=3.5cm]{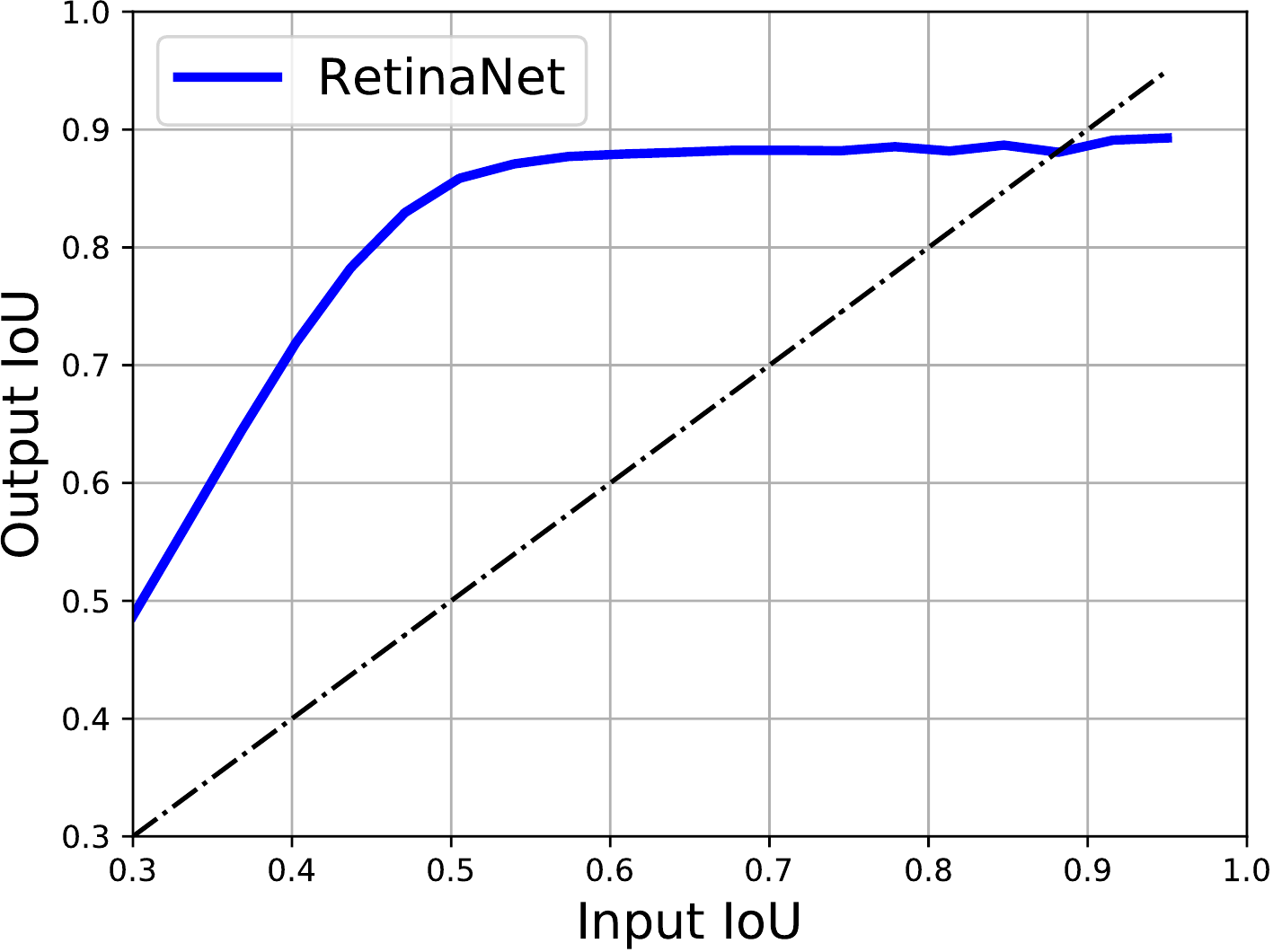}
\caption{The localization performance. The plots are based on RetinaNet with ResNet-50 backbone \cite{retinanet}.}
\label{fig:miss}
\end{figure}

Such training-inference configuration works well when the original anchor and refined anchor share the same groundtruth target. However, it fails in two circumstances. (a) When two objects are occluded to each other, the regression is apt to get confused about the moving direction. In Figure \ref{fig:demo}, there are two anchors that match the bicycle. So the detector treats these anchors as class \emph{bicycle}, then tries to optimize the classification loss and offsets between the anchors and groundtruth. However since the \emph{person} and the \emph{bicycle} are with very significant inter-class occlusion, the anchor in yellow is incorrectly regressed to the \emph{person}. The misalignment may lead to accurately located anchors being suppressed by misleading ones in the NMS procedure. (b) Some anchors assigned as negative samples may also match the groundtruth well after regression. Unfortunately these samples are eliminated due to low class scores.
In Figure \ref{fig:miss}, we plot the IoUs of the anchors with the nearest ground-truth before and after regression. Some anchors with low IoUs also match the groundtruth well after  being regressed.
Detection model that is optimized only based on the original anchors is loose and inaccurate. 
Given the above analysis, a natural question to ask is: \emph{could we use the input hypothesis of the more accurate, regressed anchors for the optimization of the detector}? Utilizing the regressed anchor hypothesis at training phase could bridge the gap between optimization and prediction for each detection. 

\begin{figure*}[t]
\centering
\subfigure[RetinaNet]{
\includegraphics[height=2.4cm]{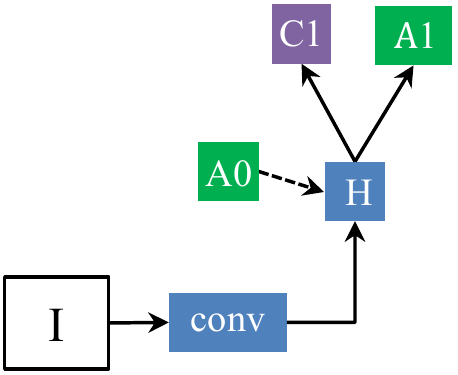}
\label{fig:choice_a}
}
\subfigure[Cascade]{
\includegraphics[height=2.4cm]{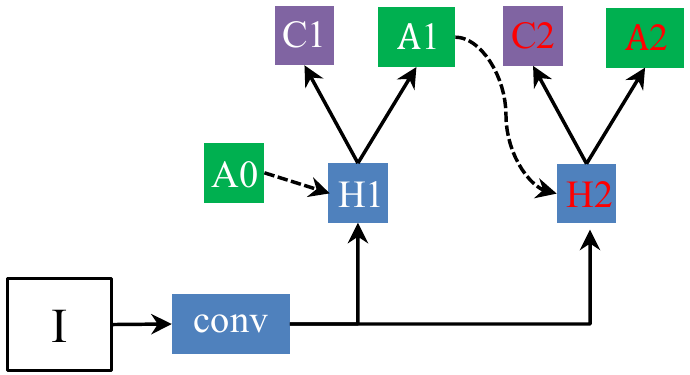}
\label{fig:choice_b}
}
\subfigure[Cascade with Gated Feature]{
\includegraphics[height=2.4cm]{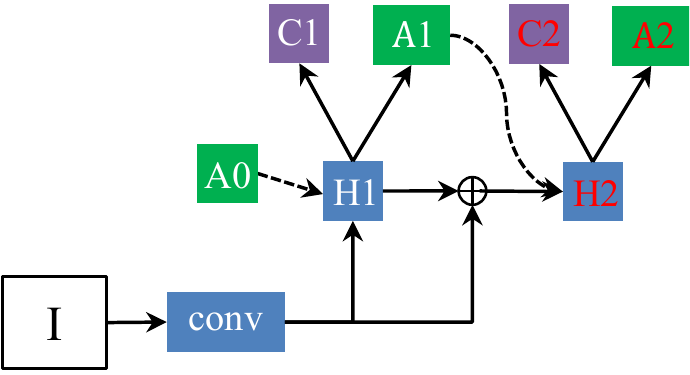}
\label{fig:choice_c}
}
\subfigure[Ours]{
\includegraphics[height=2.8cm]{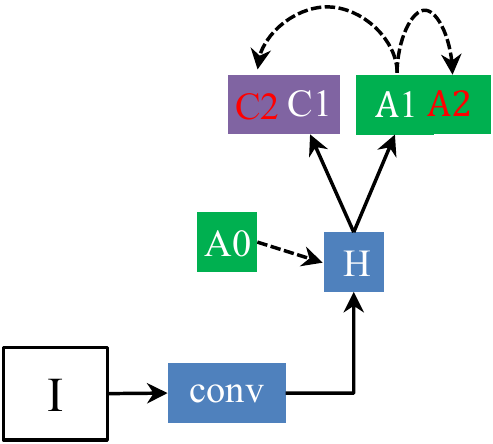}
\label{fig:choice_d}
}
\caption{The architectures of single shot detection frameworks. ``I'' is input image, ``conv'' backbone convolutions, ``H'' convolutional network head, ``$A$'' anchor box, ``C'' classification, ``$A0$'' the original anchor, and ``$A1$'' the refined anchor. In (b)-(d), the refined anchor hypothesis is utilized to further optimize the model.\protect\footnotemark}
\label{fig:choice}
\end{figure*}

There are several design choices to utilize the refined anchor for training, as shown in Figure \ref{fig:choice}. The direct way is to adopt the cascade manner, motivated from Cascade R-CNN \cite{cascadercnn}, as shown in Figure \ref{fig:choice_b} and Figure \ref{fig:choice_c}. Through comparison of several typical design choices, we find that it is the final optimization target, not the architecture designing tricks that plays the key role to boost the detector's performance. In this paper, we propose to use the refined anchor hypothesis for training the detector to keep the consistency, and to use the design architecture of Figure \ref{fig:choice_d} to implement upon the detector. In the proposed implementation, the refined anchor is used both for classification and regression. At training phase given the regular, dense sampled candidate anchors, the object detector not only adopts the original anchors to perform  classification and regression, but also utilizes the refined anchor to further optimize the same model. We name the proposed solution as consistent optimization, since it's goal is to resolve the inconsistency of training-inference for single stage object detector.

To validate it's effectiveness, we add consistent optimization on the state-of-the-art RetinaNet \cite{retinanet}. Our model, named as ConRetinaNet, is quite simple to implement and trained end-to-end.
The results show that a vanilla implementation outperforms RetinaNet with different model capacities (ResNet-50/ResNet-101), input resolutions  (short-size from 500 to 800), and localization qualities on challenging MS COCO dataset \cite{coco} (Figure \ref{fig:conretinanet}). The improvements are consistent with almost no additional computation or parameters. In particular, ConRetinaNet is able to achieve 40.1 AP on the MS COCO dataset, which is the first ResNet-101 based single stage object detector to achieve such performance without any testing time bells or whistles. We believe that this simple and effective solution can be of interest for many object detection research efforts.

\section{Related work}

\textbf{Classic Object Detectors:}
Prior to the widely development of deep
convolutional networks, the DPM \cite{dpm} and its variants \cite{fast_fp}\cite{dpm1} have been the dominating methods for years. These methods use image descriptors such as HOG \cite{hog}, SIFT \cite{sift}, and LBP \cite{lbp} as features and sweep through the entire image to find regions with a class-specific maximum response. There are also many efforts of region proposal generation. These methods usually adopt cues like superpixels \cite{ss}, edges \cite{edgebox}, saliency \cite{what_is_object} and shapes \cite{mcg} as features to generate category-independent region proposals to reduce the searching space of the target objects.

\textbf{Two-stage Detectors:} After the remarkable success of applying deep convolutional neural networks (CNNs or ConvNets) on image classification tasks, deep
learning based approaches have been actively explored for object detection, especially for the region-based convolutional neural networks (R-CNN) and its variants. The SPP-Net \cite{sppnet} and Fast R-CNN \cite{fastrcnn} speed up the R-CNN approach with RoI-Pooling that allows the classification layers to reuse the CNN feature maps. Since then, Faster R-CNN \cite{fasterrcnn} and R-FCN \cite{rfcn} replace the region proposal step with lightweight networks to deliver a complete end-to-end system.  Several attempts have been performed to boost the performance of the detector, including feature pyramid \cite{fpn}\cite{hypernet}, multi-scale \cite{snip}\cite{sniper}, and object relation \cite{relation}. The most recent related works are Cascade R-CNN \cite{cascadercnn} and IoU-Net \cite{iounet}. Cascade R-CNN gradually increases qualified proposal towards high quality detection with cascade sub-networks. IoU-Net learns to predict the IoU between each detected bounding box and the matched ground-truth  as the localization confidence.
\footnotetext{Current single stage detectors usually utilize feature pyramid heads to detect multi-scale objects. For clarity, we only show one head.}

\textbf{One-stage Detectors:} OverFeat \cite{overfeat} is one of the first modern one-stage object detector based on deep networks. SSD \cite{ssd} and YOLO \cite{yolo} have renewed interest in one-stage methods. The methods skip the region proposal generation step and predict bounding boxes and detection confidences of multiple categories directly. One stage detectors are applied over a
regular, dense sampled locations, scales, and aspect ratios, and rely on the fully ConvNets to predict the objects on each localization. The main advantage of this is its high computational efficiency \cite{yolo9000}. Due to the foreground-background class imbalance problem, previous one stage object detectors trailed in small scale object detection and larger compute budget. Recently, RetinaNet \cite{retinanet} is able to match the accuracy of the two-stage object detectors, with the proposed Focal Loss and dense prediction.

\textbf{Improving One-stage Detectors:} There are many works trying to improve the one stage object detectors, including better feature pyramid construction \cite{dssd}\cite{me_eccv}, multi-stage refinement \cite{refinedet}\cite{ssb_cascade}, adaptive anchors \cite{metaanchor} and usage of corner keypoints \cite{cornernet}. The most related works are BPN \cite{ssb_cascade} and RefineDet \cite{refinedet}. Both of them try to refine the detection box with new branches of predictions. In this work, we find that given the feature pyramid representations of an image, the key bottleneck for the performance is the train-inference misalignment, and the misalignment could be directly ameliorated by the consistent optimization. Some prior works share similarities with our work, and we will discuss them in more detail in \textsection\ref{sec:discuss}.

\section{Single Shot Object Detection}

In this section, we first review the single shot object detection. Then, we investigate the misalignment in the detector. The consistent optimization solution will be described in \textsection\ref{sec:consistent}.


\begin{figure*}[t]
\centering
\subfigure[Detection Output Mean]{
\begin{minipage}[b]{0.235\linewidth}
\includegraphics[width=1\linewidth]{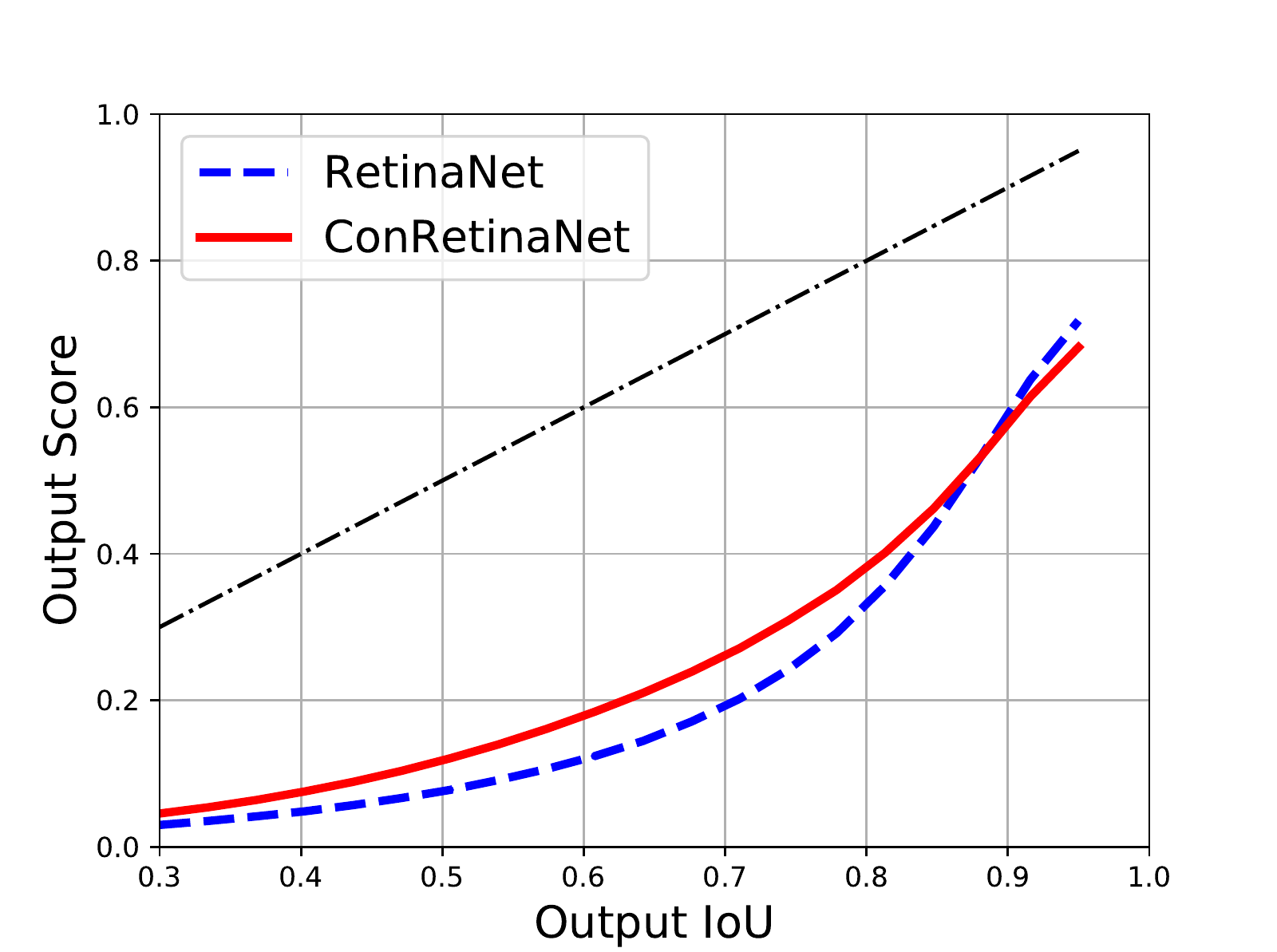}
\end{minipage}
\label{fig:conretinanet_comp_a}}
\subfigure[Detection Output Std]{
\begin{minipage}[b]{0.235\linewidth}
\includegraphics[width=1\linewidth]{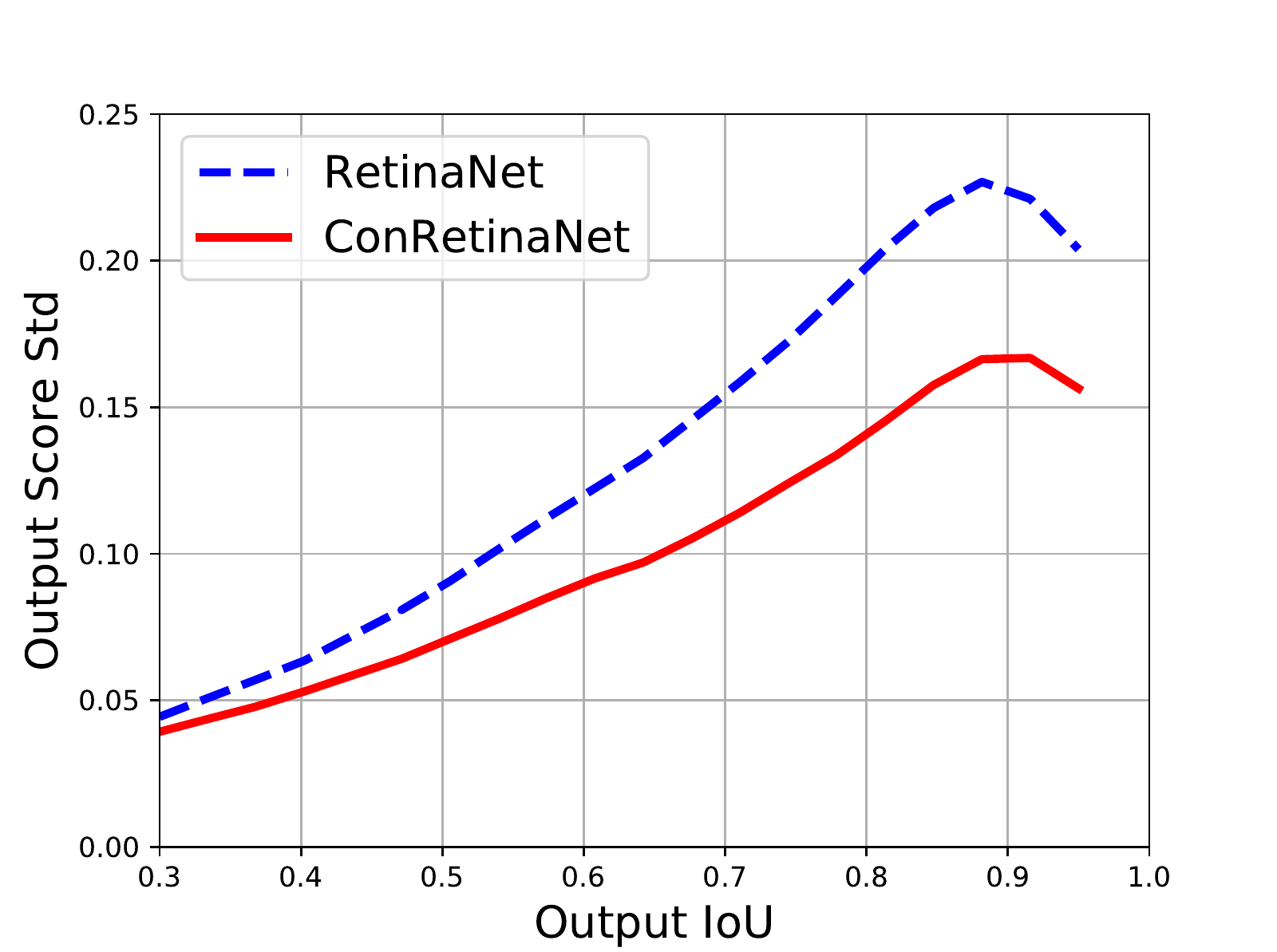}
\end{minipage}
\label{fig:conretinanet_comp_b}
}
\subfigure[Regression Output Mean]{
\begin{minipage}[b]{0.235\linewidth}
\includegraphics[width=1\linewidth]{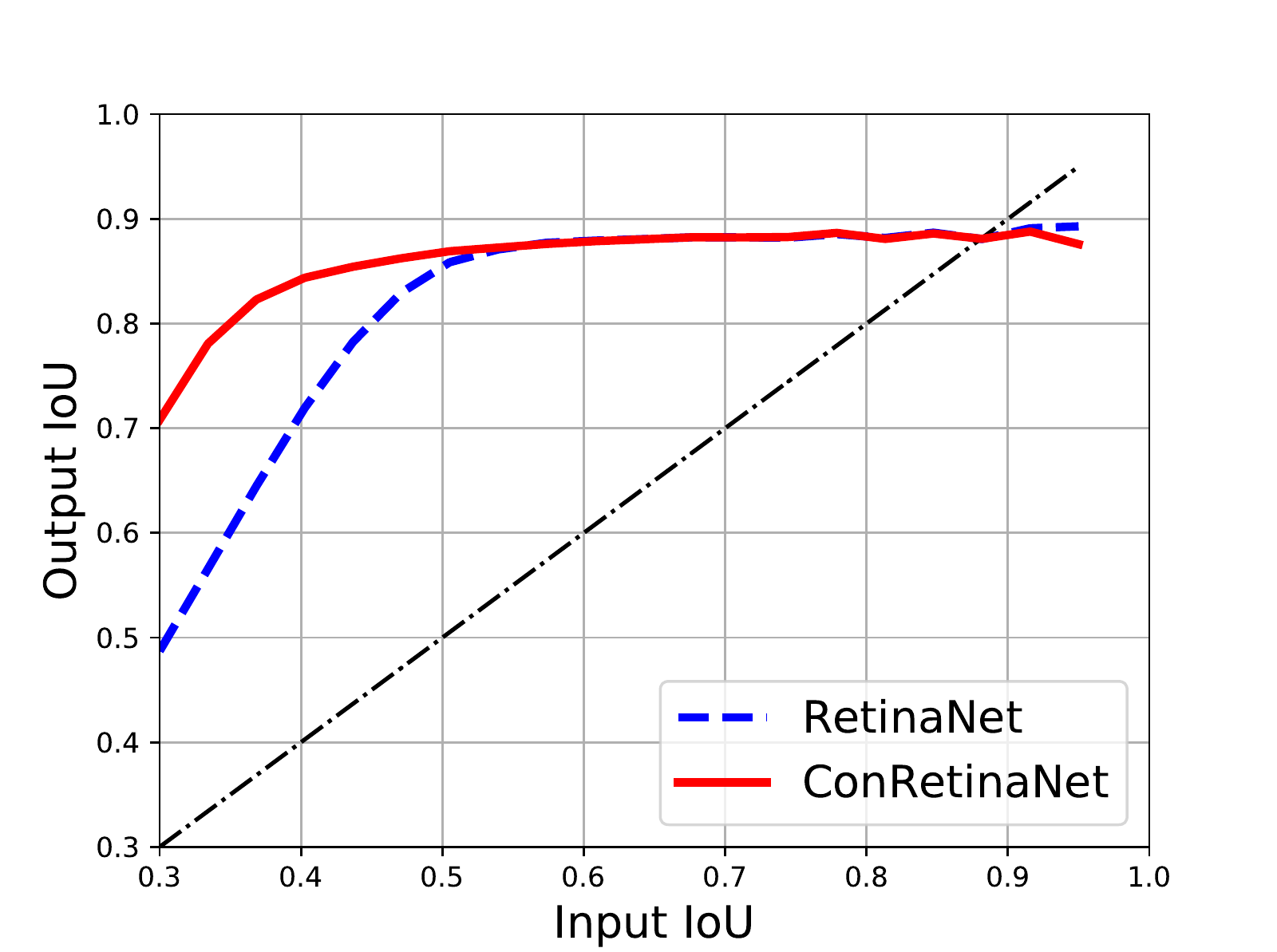}
\end{minipage}
\label{fig:conretinanet_comp_c}
}
\subfigure[Regression Output std]{
\begin{minipage}[b]{0.235\linewidth}
\includegraphics[width=1\linewidth]{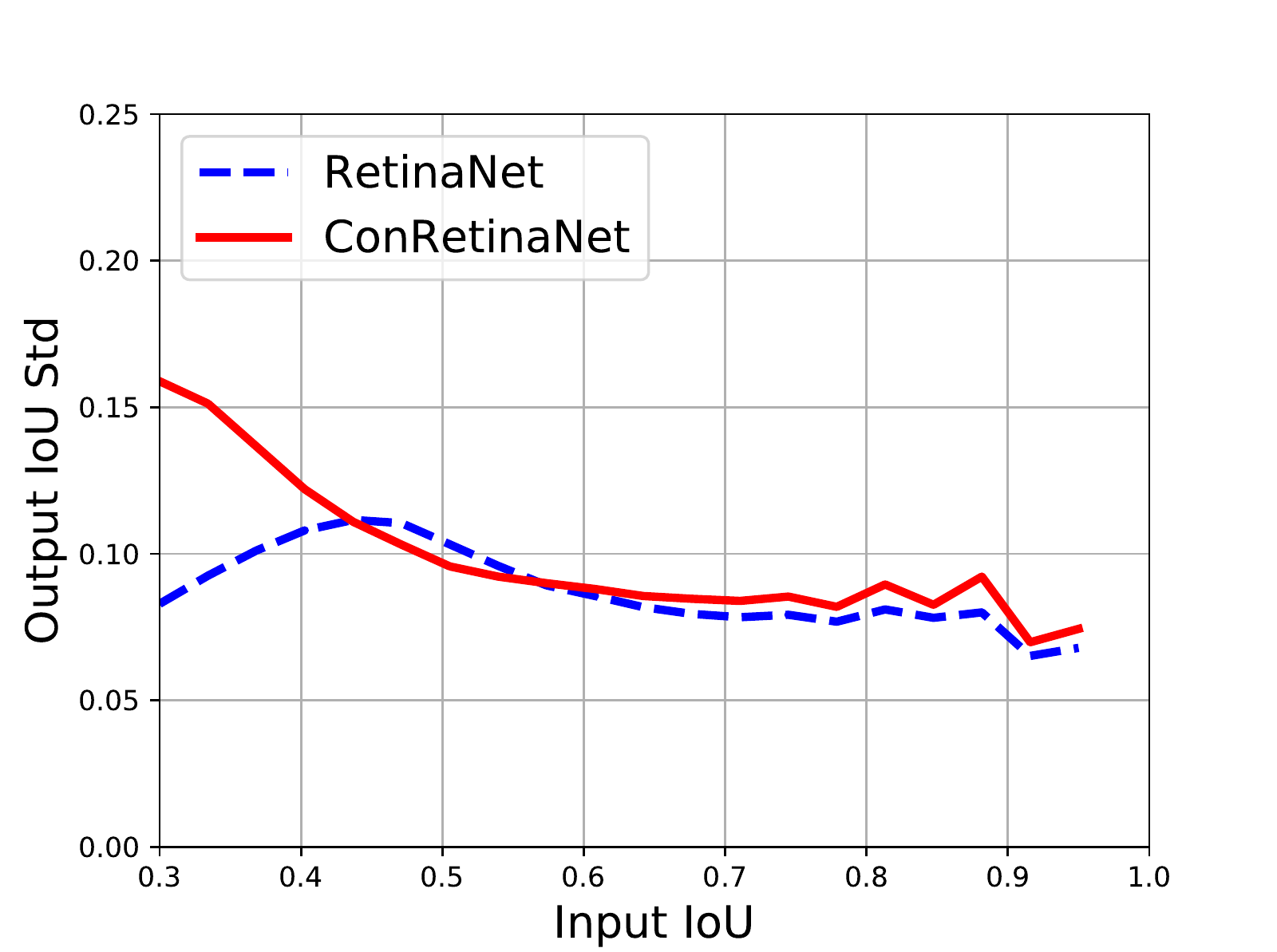}
\end{minipage}
\label{fig:conretinanet_comp_d}
}
\caption{The detection and localization performance of RetinaNet detector with different IoUs.}
\label{fig:conretinanet_comp}
\end{figure*}

The key idea of the single shot object detector is to associate the pre-defined anchor which is centered at each feature map location with the convolutional operation and results, as shown in Figure \ref{fig:choice_a}.  The single stage detector is composed of a \emph{backbone} network and two \emph{task-specific} sub-networks. The backbone is responsible for computing a convolutional feature map over an entire input image and is an off-the-shelf convolutional network. The first subnet performs convolutional object classification on the backbone's output; the second subnet performs convolutional anchor box regression. In the box sub-network, the optimization goal is to minimize the offsets between the regular anchors and the nearby ground-truth boxes, if one exists (usually utilizing the Smooth $L_1$ loss). For each of the anchors per spatial location, these 4 outputs predict the relative offset between the anchor and the ground-truth box. The classification sub-networks predict the probability of object presence at each spatial position for each of the anchors and object classes.

During inference, each pre-defined anchor is regressed based on the box sub-networks.  The score of the the refined box is given by the classification sub-networks. Finally, non-maximum-suppression (NMS) is applied to remove duplicated bounding boxes. In this paper, we take RetinaNet \cite{retinanet} to do case study since it gets promising performance both on efficiency and accuracy.

\subsection{The Misalignments}\label{sec:imbalance}

There are two misalignments between training and inference in the current single stage detector. (a) The localization qualities between original anchors and the refined anchors are very different, as shown in Figure \ref{fig:miss}. Such difference will harm the reliability of the class output scores, since the classification subnet is trained based on the original anchor. Moreover, the inter-class confusion (Figure \ref{fig:demo}) may lead to the well located box being eliminated, causing inter-class error when scoring a final bounding box. (b) Anchors whose IoU overlap with the groundtruth lower than 0.5 are treated as negative samples. However after regression, some refined anchors also have high overlaps with the groundtruth. Such samples will be eliminated due to the small class scores, causing foreground-background classification error. 

Figure \ref{fig:conretinanet_comp} shows the detection  and localization performance of RetinaNet \cite{retinanet} with different IoUs. As shown in Figure \ref{fig:conretinanet_comp_a}, the output IoUs and the prediction confidences have a significant positive correlation. However, the variations are increasing dramatically as the output IoU increases (Figure \ref{fig:conretinanet_comp_b}). That means the detector is more robust for the negative samples than the positive samples, thanks to the utilization of Focal Loss. However, the whole detection performance is evaluated as a function of IoU threshold, as in MS COCO \cite{coco}. The robustness of the positive samples is also import for object detection, since the score confidence order of the samples plays an decisive role for an accurate detector, not just filtering the negative samples. We believe the high variations are caused by the training-inference inconsistency as analyzed above.

We also visualize the localization performance of RetinaNet. The localization performance is evaluated as a function of the input IoUs. Figure \ref{fig:conretinanet_comp_c} shows the average output IoUs while Figure \ref{fig:conretinanet_comp_d} visualizes the variations.
What surprises us is that the regressor performs pretty well when the input IoUs are larger (for example, input IoU $\geq$ 0.5). One can infer that RetinaNet seems to produce more tight boxes, same as that suggested by \cite{lrp}. Utilizing the regressed anchors are more accurate for the training of a high quality object detector.

\section{Consistent Optimization}\label{sec:consistent}

\textbf{Consistent Detection:} The loose training signal for the classification sub-networks has hindered the accuracy of the detector, since the behaviors between the training anchors and refined anchors at prediction phase are different. In this work, the solution is simple: attaching subsequent classification targets for the regressed anchors. The classification target becomes
\begin{equation}
L_{cls} = \frac{1}{N_{cls}} \sum_{i} [ L_{cls} (c_i, c_i^{*}) + \alpha L_{cls} (c_i, c_i^{\dag}) ].
\label{eq:cls}
\end{equation}
Here, $i$ is the index of an anchor in a mini-batch and $c_i$ is the predicted probability of the (refined) anchor $i$ being an object. The ground-truth label $c_i^{*}$ is the label for the original anchor $i$, while $c_i^{\dag}$ is the label for the refined one. $N_{cls}$ is the mini-batch size for the classification branch. $\alpha$ balances the weights between two terms. We find that directly training the model with the refined anchors already gets superior performance. Combining the two loss together makes the training process more stable, and does not harm the performance. 

\textbf{Consistent Localization}
As shown in Figure \ref{fig:conretinanet_comp_c}, RetinaNet seems to produce tight boxes, which is different from the observations of two-stage object detectors \cite{cascadercnn}\cite{lrp}. To keep consistency with the classification branch, we also add the subsequent regressions. The localization loss function becomes
\begin{equation}
L_{reg} = \frac{1}{N^{0}_{reg}} \sum_{i} L_{reg}^{0} (t_i^{0}, t_i^{*}) +  \frac{1}{N^{1}_{reg}} \sum_{i} L_{reg}^{1} (t_i^{1}, t_i^{\dag}),
\label{eq:reg}
\end{equation}
where $t_i^{0}$ is the predicted offset of the original anchor $i$, $t_i^{1}$ the predicted offset of the refined one. $t^{*}$ and $t^{\dag}$ are corresponding groundtruch offsets for the original and refined anchor. $N_{reg}$ is the mini-batch size for the regression branch. More details about the localization loss are referred to \cite{fasterrcnn}\cite{detectron}.

\textbf{Implementation Choices:}
Given the consistent optimization functions, there are also several implementation choices. Two typical implementations are shown in Figure \ref{fig:choice_b} and Figure \ref{fig:choice_c}. In Figure \ref{fig:choice_b}, we add new sibling branches for the subsequent localization and classification, denoted as cascade version. The design of the new subnet is identical to the existing subnet, and the parameters are not shared, motivated from Cascade R-CNN \cite{cascadercnn}. In Figure \ref{fig:choice_c}, the feature head in the previous stage is also combined to enrich the feature for the subsequent prediction, denoted as gated cascade version, which is similar to some recent works that encode more semantic features for further refinement of the detector \cite{ssb_cascade}\cite{refinedet}. Finally, the proposed version is shown in Figure \ref{fig:choice_d}, which does not require more computational branches. The implementation is simple: in parallel with the original classification and localization loss, we add the consistent terms of Equation \ref{eq:cls} and \ref{eq:reg}. The input features are shared between different terms\footnote{Only the last layer's parameters of the bounding box regression are not shared.}. During inference, the confidence score are generated exactly the same way as the traditional detector.

In the experiments, we show that the implementation of Figure \ref{fig:choice_d} get comparable results compared with other variants. More importantly, it requires almost no additional computation during training and inference, and is easier to converge.

\section{Discussion and Comparison to Prior Works}\label{sec:discuss}
Here, we compare the consistent optimization with several existing object detectors, and point out key similarities and differences between them.

\textbf{Cascade R-CNN}  Cascade R-CNN \cite{cascadercnn} consists of a sequence of detectors trained with increasing IoU thresholds, to be sequentially more selective against close false positives. It's straightforward to extend the cascade idea into one stage detector. However, the main challenge is that the one stage detectors only rely on the convolutional kernels to associate the anchors and final predictions. In contrast, ROI-Pooling in region based detectors makes the feature extraction of the subsequent detector heads easier. The sample distributions between two-stage and one-stage detectors are also different.

\textbf{RefineDet}
There are same previous works trying to improve the performance of single stage detectors by utilizing several stages. In RefineDet \cite{refinedet}, the authors introduce the anchor refinement module to (1) filter out negative anchors to reduce search space for the classifier, and (2) coarsely adjust the locations. The idea is very like the the cascade manner. There are two main differences: (a) The anchor refinement module in RefineDet plays the role of the first stage (or RPN stage) in Faster R-CNN \cite{fasterrcnn}. It predicts a coarse location for each anchor. Our solution is to make the final prediction more reliable by utilizing consistent optimization. (b) The RefineDet relies on a transfer connection block to transfer the features in the anchor refinement module to the object detection module. From our observation, adding more parameters is not necessary under standard feature pyramid networks \cite{fpn}. The main performance bottleneck is the misalignment between optimization and prediction. We also conduct experiments to verify this assumption.

\textbf{IoU-Net}
Recently, IoU-Net \cite{iounet} proposes to learn the IoU between the predicted bounding box and
the ground truth bounding box. IoU-NMS is then applied to the detection boxes, guided by the learned IoU. The goal of IoU-Net is to make the confidence score of the box to be consistent with the localization performance. IoU-Net shows its effectiveness on the two-stage detectors via jittered RoI training. We believe it could also be utilized in the one-stage detectors, which is beyond the scope of this paper.

\section{Experiments}\label{sec:exp}
We present experimental results on the bounding box detection track of the challenging MS COCO benchmark \cite{coco}. For training, we follow common practice \cite{detectron} and use
the MS COCO \texttt{trainval35k} split (union of 80k images from \texttt{train} and a random 35k subset of images from the 40k image \texttt{val} split). If not specified, we report studies by evaluating on the \texttt{minival5k} split. The COCO-style Average Precision (AP) averages AP across IoU thresholds from 0.5 to 0.95 with an interval of 0.05. These metrics measure the detection performance of various qualities. Final results are also reported on the \texttt{test-dev} set.

\subsection{Implementation Details}\label{sec:imp}
All experiments are implemented with Caffe2 on Detectron \cite{detectron} codebase for fair comparison. End-to-end training is used for simplicity. We replace the traditional classification and localization loss function with Equation \ref{eq:cls} and \ref{eq:reg}, unless otherwise noted. The optimization targets are all based on the refined anchors after the anchor regression. We set $\alpha = 1$ and it works well for all experiments. During training, no data augmentation is used except standard horizontal image flipping . Inference is performed on a single image scale, with no further bells and whistles. 

\subsection{Baseline}
To valid the effectiveness of the consistent optimization, we conduct experiments based on the most recently proposed RetinaNet \cite{retinanet}. RetinaNet is able to match the speed of previous one-stage detectors while get comparable accuracy  compared with existing state-of-the-art two-stage detectors. We use data parallel synchronized SGD over 4 GPUs with a total of 8 images per mini-batch (2 images per GPU). Unless otherwise specified, the experimental settings are exactly the same as that in \cite{detectron}. For all ablation studies we use an image scale with short side 500 pixels for training and testing using ResNet-50 with a Feature Pyramid Network (FPN) \cite{fpn} constructed on top.  More model setting details are referred to \cite{retinanet}. We conduct ablation studies and analyze the behavior of the consistent optimization with various design choices.

\subsection{Ablation Study}
\textbf{Comparison with different design choices:}  We first compare the design choices of the consistent optimization. The cascade and gated cascade version can be seen as cascade extensions of RetinaNet. Table \ref{table:choice} shows the performances on MS COCO dataset. To our surprise, using new features does not help boosting the performance, even we combine the previous prediction head. The last implementation in Figure \ref{fig:choice} enjoys both accuracy gains and efficiency.
\begin{table}[h]
\begin{center}
\begin{tabular}{l|cccc}
 & AP & AP$_{50}$ & AP$_{75}$\\
\hline\hline
RetinaNet baseline & 32.5& 50.9& 34.8 \\
\hline
+ Cascade  & 33.4& 51.3& 36.0 \\
+ Gated Cascade & 33.6& 51.5& 36.3\\
\hline
+ Consistent Only & 33.7& 51.7& 36.2 \\
\end{tabular}
\end{center}
\caption{The impact of the different design choices.}
\label{table:choice}
\end{table}

In the cascade setting, we rely the new branches to fit the data after the regression, like Cascade R-CNN \cite{cascadercnn} does. However, the position of anchor box has been changed after the first regression. In Cascade R-CNN, the authors utilize the ROI-Pooling to extract more accurate features. In the single stage detectors, it is hard for the convolutional networks to learn such transformations in-place. We also tried to use the deformable convolution \cite{dcn} to learn the transformations, but failed. This experiment demonstrates that the improvements come from better training of the detector, not more parameters or architecture designs.

\textbf{The Hyper-parameters:} The hyper-parameters in the consistent terms are the IoU thresholds $\mu_{pos}$ and $\mu_{neg}$, which define the positive and negative samples. We conduct several experiments to compare the sensitivity of the hyper-parameters, as shown in Table \ref{table:hyperparam}. The detection performance of the model is robust to the hyper-parameters. Higher $\mu_{pos}$ gets slight better performance on strict IoU thresholds. However, the whole performances are similar. We use the setting of $\mu_{pos}=0.6$ and $\mu_{neg}=0.5$ for all other experiments in this work.

\begin{table}[h]
\begin{center}
\begin{tabular}{p{0.35cm}p{0.54cm}|cp{0.5cm}p{0.5cm}p{0.5cm}p{0.5cm}p{0.5cm}}
$\mu_{pos}$ &$\mu_{neg}$& AP & AP$_{50}$ & AP$_{60}$& AP$_{70}$ & AP$_{80} $ & AP$_{90}$\\
\hline\hline
0.5 & 0.5 & 33.5& 51.7& 47.5 & 40.7 & 29.4 & 11.2 \\
0.6 & 0.5 & 33.7& 51.7& 47.6 & 41.0 & 30.0 & 13.6 \\
0.7 & 0.6 & 33.6& 51.4& 47.6 & 40.9 & 30.2 & 14.0 \\
\end{tabular}
\end{center}
\caption{The impact of the IoU threshold $\mu_{pos}$ and $\mu_{neg}$ on detection performance.}
\label{table:hyperparam}
\end{table}

\textbf{More Stages?}
Like Cascade R-CNN, we add more stages of classifications and localizations to compare the influence. The impact of the number of stages is summarized in Table \ref{table:terms}. Adding consistent classification loss significantly improves the baseline detector (+0.8 AP). Two classification and regression terms get the best performance. Adding more regression stages leads to a slight performance decrease, while including more classification terms leads to no improvement. When we perform anchor regression once using the model trained by consistent classification and localization, the improvements are also significant (+1.0 AP). That means the consistent localization helps training the original anchor to predict the groundtruth location.

\begin{table}[h]
\begin{center}
\begin{tabular}{p{0.3cm}p{0.5cm}|cp{0.5cm}p{0.5cm}p{0.5cm}p{0.5cm}p{0.5cm}}
\#cls&\#reg& AP & AP$_{50}$ & AP$_{60}$& AP$_{70}$ & AP$_{80} $ & AP$_{90}$\\
\hline\hline
1 & 1 & 32.5& 50.6& 46.4 & 39.5 & 28.6 & 10.4 \\
\hline
2 & 1 & 33.3& 51.2& 47.4 & 40.6 & 28.9 & 11.2 \\
2 & 2$^*$ & 33.5& 51.3& 47.4 & 40.8 & 29.8 & 11.8 \\
2 & 2 & 33.7& 51.7& 47.6 & 41.0 & 30.0 & 13.6 \\
3 & 2 & 33.6& 51.7& 47.7 & 41.0 & 29.6 & 13.4 \\
\end{tabular}
\end{center}
\caption{The impact of the number of stages on detection performance. $\mu_{pos}^{2}=0.6$, $\mu_{neg}^{2}=0.5$, $\mu_{pos}^{3}=0.7$, $\mu_{neg}^{3}=0.6$ in this experiment setting. The first row is the baseline model result. ``*'' means performing anchor regression once during inference.}
\label{table:terms}
\end{table}

Both consistent classification and localization improve the performance at different IoU thresholds. The gap of 1.2 AP, shows the effectiveness of the consistent optimization for training the single shot detector. Different from previous works that always perform better for strict IoU thresholds \cite{cascadercnn}\cite{iounet}, consistent optimization enjoys gains for all localization qualities.

\begin{table*}[t]
\begin{center}
\begin{tabular}{c|c|c|cccccc}
backbone &scale& consistent &AP & AP$_{50}$ & AP$_{75}$&AP$_{S}$ & AP$_{M}$ & AP$_{L}$\\
\hline\hline
ResNet-50 & 500 & \ding{55} & 32.5 & 50.9 & 34.8 & 13.9 & 35.8 & 46.7 \\
ResNet-50 & 500 & \ding{51} & 33.7 & 51.7 & 36.2 & 14.6 & 37.7 & 49.6 \\
\hline
ResNet-50 & 800 & \ding{55} & 35.7 & 55.0 & 38.5 & 18.9 & 38.9 & 46.3 \\
ResNet-50 & 800 & \ding{51} & 36.4 & 55.4 & 39.2 & 20.0 & 39.6 & 49.7 \\
\hline
ResNet-101 & 500 & \ding{55} & 34.4 & 53.1 & 36.8 & 14.7 & 38.8 & 49.1 \\
ResNet-101 & 500 & \ding{51} & 35.5 & 53.5 & 38.3 & 16.2 & 39.8 & 52.0 \\
\hline
ResNet-101 & 800 & \ding{55} & 37.8 & 57.5 & 40.8 & 20.2 & 41.1 & 49.2 \\
ResNet-101 & 800 & \ding{51} & 38.7 & 58.2 & 41.7 & 21.4 & 42.6 & 52.2 \\
\end{tabular}
\end{center}
\caption{Detailed comparison on different resolutions and model capacities.}
\label{table:scale_depth}
\end{table*}

\textbf{Parameter and Timing}:
The number of the parameters are almost not increased with consistent optimization, at both training and inference phases. During training the consistent loss needs to compute the classification and localization targets of the refined boxes. At inference, we only add the two-stage regression which is implemented by a single convolutional layer. Recent works on boosting one stage detectors \cite{refinedet}\cite{ssb_cascade} or two stage detectors \cite{relation}\cite{iounet}\cite{cascadercnn} mostly rely on more parameters compared with the baseline models.


\subsection{Generalization Capacity}
\textbf{Across model depth and scale:}
To validate the generalization capacity of consistent optimization, we conduct experiments across model depth and input scale based on RetinaNet \cite{retinanet}. As shown in Table \ref{table:scale_depth}, the proposed method  improves on these baselines consistently by $\sim$1.0 point, independently of model capacities and input scales. These results suggest that the consistent optimization is widely applicable within one stage object detector. When analysing the performance on small, medium and large object scales, we observe that most improvements come from the larger objects.

\textbf{Results on SSD:}
We further do experiments under Single Shot MultiBox Detector (SSD) \cite{ssd} baseline to validate its generalization capacity. SSD is built on top of a ``base'' network that ends with some convolutional layers. Each of the added layers, and some of the earlier base network layers are used to predict scores and offsets for the predefined anchor boxes. The experiment is conducted based on re-implementation of SSD512 using PyTorch \cite{pytorch}, more details are referred to \cite{ssd}. We use the VGG-16 \cite{vgg} and ResNet-50 \cite{resnet} models pretrained on the \texttt{ImageNet1k} as the start models\footnote{\url{https://github.com/pytorch/vision}}, and fine-tune them on the MS COCO dataset.
\begin{table}[h]
\begin{center}
\begin{tabular}{c|c|ccc}
backbone&consistent& AP & AP$_{50}$ & AP$_{75}$\\
\hline\hline
VGG-16 \cite{ssd}& \ding{55} & 28.8& 48.5& 30.3 \\
VGG-16& \ding{55} & 28.9& 47.9& 30.6 \\
VGG-16& \ding{51} & 30.5& 49.6& 31.7 \\
\hline
ResNet-50& \ding{55} & 30.6& 50.0& 32.2 \\
ResNet-50& \ding{51} & 31.9& 51.0& 33.8 \\
\end{tabular}
\end{center}
\caption{The impact of the consistent optimization on SSD detector.}
\label{table:ssd}
\end{table}

Table \ref{table:ssd} shows the comparison results when adding consistent loss on SSD baseline. The first row is the results  reported in the paper. Others are our re-implementation results using PyTorch.
The detection results show that the consistent optimization also has significant improvements over the popular SSD architecture. These reinforce our belief on the generalization capacity of the consistent optimization.

\begin{table*}[t]
\begin{center}
\begin{tabular}{p{4.3cm}|c|cccccc}
& backbone &AP & AP$_{50}$ & AP$_{75}$&AP$_{S}$ & AP$_{M}$ & AP$_{L}$\\
\hline\hline
\emph{two-stage}     &&&&&&&\\
~~Faster R-CNN+++\cite{resnet}*     &ResNet-101 & 34.9 & 55.7 & 37.4 & 15.6 & 38.7 & 50.9 \\
~~Faster R-CNN by G-RMI\cite{grmi} &Inception-ResNet-v2&34.7 &55.5 &36.7 &13.5 &38.1 &52.0\\
~~Faster R-CNN w FPN\cite{fpn}  &ResNet-101 & 36.2 & 59.1 & 39.0 & 18.2 & 39.0 & 48.2 \\
~~Faster R-CNN w TDM\cite{tdm}  &Inception-ResNet-v2  & 36.8 & 57.7 & 39.2 & 16.2 & 39.8 & 52.1 \\
~~Deformable R-FCN \cite{dcn}* &Aligned-Inception-ResNet &37.5& 58.0 &40.8 &19.4 &40.1 &52.5 \\
~~Mask R-CNN\cite{maskrcnn}         &ResNet-101 & 38.2 & 60.3 & 41.7 & 20.1 & 41.1 & 50.2 \\
~~Relation\cite{relation}           &DCN-101    & 39.0 & 58.6 & 42.9 & - & - & - \\
~~Regionlets \cite{relation}           &ResNet-101    & 39.3 & 59.8 & - & 21.7 & 43.7 & 50.9 \\
~~DeNet768 \cite{denet}      &ResNet-101 & 39.5 &58.0 &42.6 &18.9 &43.5 &54.1 \\
~~IoU-Net\cite{iounet}          &ResNet-101 & 40.6 & 59.0 & - & - & - & - \\
~~Cascade R-CNN\cite{cascadercnn}       &ResNet-101 & 42.8 & 62.1 & 46.3 & 23.7 & 45.5 & 55.2 \\
~~Faster R-CNN by MSRA\cite{dcnv2}       &DCN-v2-101 & 44.0 &65.9 &48.1 &23.2 &47.7 &59.6 \\
\hline
\emph{one-stage}     &&&&&&&\\
~~YOLOv2 \cite{yolo}             &DarkNet-19 & 21.6 & 44.0 & 19.2 & 5.0 & 22.4 & 35.5 \\
~~RON384 \cite{ron}*            &VGG-16 & 27.4 & 49.5 & 27.1 & - & - & - \\
~~SSD513 \cite{dssd}              &ResNet-101 & 31.2 & 50.4 & 33.3 & 10.2 & 34.5 & 49.8 \\
~~YOLOv3 \cite{yolov3}            &Darknet-53 & 33.0 &  57.9 &  34.4 & 18.3 & 35.4 & 41.9 \\
~~DSSD513 \cite{dssd}            &ResNet-101 & 33.2 & 53.3 & 35.2 & 13.0 & 35.4 & 51.1 \\
~~RefineDet512 \cite{refinedet}      &ResNet-101 & 36.4 & 57.5 & 39.5 & 16.6 & 39.9 & 51.4 \\
~~RetinaNet \cite{retinanet}          &ResNet-101 & 39.1 & 59.1 & 42.3 & 21.8 & 42.7 & 50.2 \\
~~CornerNet511\cite{cornernet}*        &Hourglass-104 & 40.5 & 56.5 & 43.1 & 19.4 & 42.7 & 53.9 \\
\hline
\emph{ours}     &&&&&&&\\
~~ConRetinaNet        &ResNet-50 & 37.3 & 56.4 & 40.3 & 21.3 & 40.2 & 51.0 \\
~~ConRetinaNet        &ResNet-101 & 40.1 & 59.6 & 43.5 & 23.4 & 44.2 & 53.3 \\

\end{tabular}
\end{center}
\caption{Object detection single-model results (bounding box AP) \emph{v.s.} state-of-the-art on COCO \texttt{test-dev}. We show results for our
ConRetinaNet-50 and ConRetinaNet-101 models with 800 input scale. Both RetinaNet and ConRetinaNet are trained with scale jitter and for 1.5$\times$ longer than the same model from Table \ref{table:scale_depth}. Our model achieves top results,
outperforming most one-stage and two-stage models.  The entries denoted by ``*'' used bells and whistles at inference.}
\label{table:sota}
\end{table*}

\subsection{Comparison to State of the Art}
The consistent optimization extension of RetinaNet, is compared to state-of-the-art object detectors (both one-stage and two-stage) in Table \ref{table:sota}.  We report the standard COCO metrics including AP (averaged over IoU thresholds), AP$_{50}$, AP$_{75}$, and AP$_S$, AP$_M$, AP$_L$ (AP at different scales) on the \texttt{test-dev} set. The experimental settings are described in \textsection\ref{sec:imp}.

The first group of detectors on Table \ref{table:sota} are two-stage detectors, the second group one-stage detectors, and the last group the consistent optimization extension of RetinaNet. The extension from RetinaNet to ConRetinaNet improves detection performance by $\sim$1 point, and it also \textbf{outperforms all single-stage detector under ResNet-101 backbone, under all evaluation metrics}. This includes the very recent one-stage RefineDet \cite{refinedet} and two-stage relation networks \cite{relation}. Specifically, ConRetinaNet-ResNet-50 outperforms DSSD-ResNet-101 \cite{dssd} and RefineDet-ResNet-101 \cite{refinedet} with large margins. We also compare the single model ConRetinaNet with CornerNet \cite{cornernet}, which uses heavier Hourglass-104 and more training and testing time data augmentations.

We note that the Cascade R-CNN \cite{cascadercnn} and Faster R-CNN based on Deformbale ConvNets v2 \cite{dcnv2} show better accuracy than ConRetinaNet. The difficulty to bring the cascade idea to the single stage detectors is how to associate the refined anchors with the corresponding features. An intersecting direction is to utilize the region based branch to get more accurate features. The deformable convolution shows better capability to model the geometric transformation of the objects. Replacing the backbone networks with Deformable ConvNets is supposed to get better performance, which is beyond the focus of this paper. At last, Cascade R-CNN and Deformbale ConvNets both require more parameters to get such results.

\section{Conclusion and Future Work}
In this paper, we propose the simple and  effective consistent optimization to boost the performance of single stage object detectors. By examination of the model behaviors, we find that the optimization misalignment between training and inference is the bottleneck to get better results. We conduct extensive experiments to compare different model design choices, and demonstrate that the consistent optimization is the most important factor. Utilizing consistent optimization requires almost no additional parameters, and it shows its effectiveness using the strong RetinaNet baseline on challenging MS COCO dataset.

For the future work, we will try to combine the localization confidence which is proposed in \cite{iounet} and  consistent optimization to further improve the detector's quality. Another important direction is to associate the refined boxes and their corresponding features with geometric transformation on the feature maps.

{\small
\bibliographystyle{ieee}
\bibliography{egbib}

\begin{thebibliography}{10}\itemsep=-1pt

\bibitem{what_is_object}
B.~Alexe, T.~Deselaers, and V.~Ferrari.
\newblock What is an object?
\newblock In {\em Computer Vision and Pattern Recognition (CVPR), 2010 IEEE
  Conference on}, pages 73--80. IEEE, 2010.

\bibitem{mcg}
P.~Arbel{\'a}ez, J.~Pont-Tuset, J.~T. Barron, F.~Marques, and J.~Malik.
\newblock Multiscale combinatorial grouping.
\newblock In {\em Proceedings of the IEEE conference on computer vision and
  pattern recognition}, pages 328--335, 2014.

\bibitem{dpm1}
H.~Azizpour and I.~Laptev.
\newblock Object detection using strongly-supervised deformable part models.
\newblock In {\em European Conference on Computer Vision}, pages 836--849.
  Springer, 2012.

\bibitem{cascadercnn}
Z.~Cai and N.~Vasconcelos.
\newblock Cascade r-cnn: Delving into high quality object detection.
\newblock {\em arXiv preprint arXiv:1712.00726}, 2017.

\bibitem{rfcn}
J.~Dai, Y.~Li, K.~He, and J.~Sun.
\newblock R-fcn: Object detection via region-based fully convolutional
  networks.
\newblock In {\em Advances in neural information processing systems}, pages
  379--387, 2016.

\bibitem{dcn}
J.~Dai, H.~Qi, Y.~Xiong, Y.~Li, G.~Zhang, H.~Hu, and Y.~Wei.
\newblock Deformable convolutional networks.
\newblock In {\em Proceedings of the IEEE International Conference on Computer
  Vision}, pages 764--773, 2017.

\bibitem{hog}
N.~Dalal and B.~Triggs.
\newblock Histograms of oriented gradients for human detection.
\newblock In {\em Computer Vision and Pattern Recognition, 2005. CVPR 2005.
  IEEE Computer Society Conference on}, volume~1, pages 886--893. IEEE, 2005.

\bibitem{fast_fp}
P.~Doll{\'a}r, R.~Appel, S.~Belongie, and P.~Perona.
\newblock Fast feature pyramids for object detection.
\newblock {\em IEEE Transactions on Pattern Analysis and Machine Intelligence},
  36(8):1532--1545, 2014.

\bibitem{dpm}
P.~F. Felzenszwalb, R.~B. Girshick, D.~McAllester, and D.~Ramanan.
\newblock Object detection with discriminatively trained part-based models.
\newblock {\em IEEE transactions on pattern analysis and machine intelligence},
  32(9):1627--1645, 2010.

\bibitem{dssd}
C.-Y. Fu, W.~Liu, A.~Ranga, A.~Tyagi, and A.~C. Berg.
\newblock Dssd: Deconvolutional single shot detector.
\newblock {\em arXiv preprint arXiv:1701.06659}, 2017.

\bibitem{fastrcnn}
R.~Girshick.
\newblock Fast r-cnn.
\newblock In {\em Proceedings of the IEEE international conference on computer
  vision}, pages 1440--1448, 2015.

\bibitem{detectron}
R.~Girshick, I.~Radosavovic, G.~Gkioxari, P.~Doll\'{a}r, and K.~He.
\newblock Detectron.
\newblock \url{https://github.com/facebookresearch/detectron}, 2018.

\bibitem{maskrcnn}
K.~He, G.~Gkioxari, P.~Doll{\'a}r, and R.~Girshick.
\newblock Mask r-cnn.
\newblock In {\em Computer Vision (ICCV), 2017 IEEE International Conference
  on}, pages 2980--2988. IEEE, 2017.

\bibitem{sppnet}
K.~He, X.~Zhang, S.~Ren, and J.~Sun.
\newblock Spatial pyramid pooling in deep convolutional networks for visual
  recognition.
\newblock In {\em European conference on computer vision}, pages 346--361.
  Springer, 2014.

\bibitem{resnet}
K.~He, X.~Zhang, S.~Ren, and J.~Sun.
\newblock Deep residual learning for image recognition.
\newblock In {\em Proceedings of the IEEE conference on computer vision and
  pattern recognition}, pages 770--778, 2016.

\bibitem{relation}
H.~Hu, J.~Gu, Z.~Zhang, J.~Dai, and Y.~Wei.
\newblock Relation networks for object detection.
\newblock In {\em Computer Vision and Pattern Recognition (CVPR)}, volume~2,
  2018.

\bibitem{grmi}
J.~Huang, V.~Rathod, C.~Sun, M.~Zhu, A.~Korattikara, A.~Fathi, I.~Fischer,
  Z.~Wojna, Y.~Song, S.~Guadarrama, et~al.
\newblock Speed/accuracy trade-offs for modern convolutional object detectors.
\newblock In {\em IEEE CVPR}, volume~4, 2017.

\bibitem{iounet}
B.~Jiang, R.~Luo, J.~Mao, T.~Xiao, and Y.~Jiang.
\newblock Acquisition of localization confidence for accurate object detection.
\newblock In {\em Proceedings of the European Conference on Computer Vision,
  Munich, Germany}, pages 8--14, 2018.

\bibitem{me_eccv}
T.~Kong, F.~Sun, W.~Huang, and H.~Liu.
\newblock Deep feature pyramid reconfiguration for object detection.
\newblock In {\em European Conference on Computer Vision}, pages 172--188.
  Springer, 2018.

\bibitem{ron}
T.~Kong, F.~Sun, A.~Yao, H.~Liu, M.~Lu, and Y.~Chen.
\newblock Ron: Reverse connection with objectness prior networks for object
  detection.
\newblock In {\em IEEE Conference on Computer Vision and Pattern Recognition},
  volume~1, page~2, 2017.

\bibitem{hypernet}
T.~Kong, A.~Yao, Y.~Chen, and F.~Sun.
\newblock Hypernet: Towards accurate region proposal generation and joint
  object detection.
\newblock In {\em Proceedings of the IEEE conference on computer vision and
  pattern recognition}, pages 845--853, 2016.

\bibitem{cornernet}
H.~Law and J.~Deng.
\newblock Cornernet: Detecting objects as paired keypoints.
\newblock In {\em Proceedings of the European Conference on Computer Vision
  (ECCV)}, volume~6, 2018.

\bibitem{fpn}
T.-Y. Lin, P.~Doll{\'a}r, R.~B. Girshick, K.~He, B.~Hariharan, and S.~J.
  Belongie.
\newblock Feature pyramid networks for object detection.
\newblock In {\em CVPR}, volume~1, page~4, 2017.

\bibitem{retinanet}
T.-Y. Lin, P.~Goyal, R.~Girshick, K.~He, and P.~Doll{\'a}r.
\newblock Focal loss for dense object detection.
\newblock {\em IEEE transactions on pattern analysis and machine intelligence},
  2018.

\bibitem{coco}
T.-Y. Lin, M.~Maire, S.~Belongie, J.~Hays, P.~Perona, D.~Ramanan,
  P.~Doll{\'a}r, and C.~L. Zitnick.
\newblock Microsoft coco: Common objects in context.
\newblock In {\em European conference on computer vision}, pages 740--755.
  Springer, 2014.

\bibitem{ssd}
W.~Liu, D.~Anguelov, D.~Erhan, C.~Szegedy, S.~Reed, C.-Y. Fu, and A.~C. Berg.
\newblock Ssd: Single shot multibox detector.
\newblock In {\em European conference on computer vision}, pages 21--37.
  Springer, 2016.

\bibitem{sift}
D.~G. Lowe.
\newblock Distinctive image features from scale-invariant keypoints.
\newblock {\em International journal of computer vision}, 60(2):91--110, 2004.

\bibitem{sniper}
M.~Najibi, B.~Singh, and L.~S. Davis.
\newblock Autofocus: Efficient multi-scale inference.
\newblock {\em arXiv preprint arXiv:1812.01600}, 2018.

\bibitem{lrp}
K.~Oksuz, B.~C. Cam, E.~Akbas, and S.~Kalkan.
\newblock Localization recall precision (lrp): A new performance metric for
  object detection.
\newblock In {\em European Conference on Computer Vision (ECCV)}, volume~6,
  2018.

\bibitem{pytorch}
A.~Paszke, S.~Gross, S.~Chintala, G.~Chanan, E.~Yang, Z.~DeVito, Z.~Lin,
  A.~Desmaison, L.~Antiga, and A.~Lerer.
\newblock Automatic differentiation in pytorch.
\newblock 2017.

\bibitem{yolo}
J.~Redmon, S.~Divvala, R.~Girshick, and A.~Farhadi.
\newblock You only look once: Unified, real-time object detection.
\newblock In {\em Proceedings of the IEEE conference on computer vision and
  pattern recognition}, pages 779--788, 2016.

\bibitem{yolo9000}
J.~Redmon and A.~Farhadi.
\newblock Yolo9000: better, faster, stronger.
\newblock {\em arXiv preprint}, 2017.

\bibitem{yolov3}
J.~Redmon and A.~Farhadi.
\newblock Yolov3: An incremental improvement.
\newblock {\em arXiv preprint arXiv:1804.02767}, 2018.

\bibitem{fasterrcnn}
S.~Ren, K.~He, R.~Girshick, and J.~Sun.
\newblock Faster r-cnn: Towards real-time object detection with region proposal
  networks.
\newblock In {\em Advances in neural information processing systems}, pages
  91--99, 2015.

\bibitem{overfeat}
P.~Sermanet, D.~Eigen, X.~Zhang, M.~Mathieu, R.~Fergus, and Y.~LeCun.
\newblock Overfeat: Integrated recognition, localization and detection using
  convolutional networks.
\newblock {\em arXiv preprint arXiv:1312.6229}, 2013.

\bibitem{tdm}
A.~Shrivastava, R.~Sukthankar, J.~Malik, and A.~Gupta.
\newblock Beyond skip connections: Top-down modulation for object detection.
\newblock {\em arXiv preprint arXiv:1612.06851}, 2016.

\bibitem{vgg}
K.~Simonyan and A.~Zisserman.
\newblock Very deep convolutional networks for large-scale image recognition.
\newblock {\em arXiv preprint arXiv:1409.1556}, 2014.

\bibitem{snip}
B.~Singh and L.~S. Davis.
\newblock An analysis of scale invariance in object detection--snip.
\newblock In {\em Proceedings of the IEEE Conference on Computer Vision and
  Pattern Recognition}, pages 3578--3587, 2018.

\bibitem{denet}
L.~Tychsen-Smith and L.~Petersson.
\newblock Improving object localization with fitness nms and bounded iou loss.
\newblock {\em arXiv preprint arXiv:1711.00164}, 2017.

\bibitem{ss}
K.~E. Van~de Sande, J.~R. Uijlings, T.~Gevers, and A.~W. Smeulders.
\newblock Segmentation as selective search for object recognition.
\newblock In {\em Computer Vision (ICCV), 2011 IEEE International Conference
  on}, pages 1879--1886. IEEE, 2011.

\bibitem{lbp}
X.~Wang, T.~X. Han, and S.~Yan.
\newblock An hog-lbp human detector with partial occlusion handling.
\newblock In {\em Computer Vision, 2009 IEEE 12th International Conference on},
  pages 32--39. IEEE, 2009.

\bibitem{ssb_cascade}
X.~Wu, D.~Zhang, J.~Zhu, and S.~C. Hoi.
\newblock Single-shot bidirectional pyramid networks for high-quality object
  detection.
\newblock {\em arXiv preprint arXiv:1803.08208}, 2018.

\bibitem{metaanchor}
T.~Yang, X.~Zhang, Z.~Li, W.~Zhang, and J.~Sun.
\newblock Metaanchor: Learning to detect objects with customized anchors.
\newblock In {\em Advances in Neural Information Processing Systems}, pages
  318--328, 2018.

\bibitem{refinedet}
S.~Zhang, L.~Wen, X.~Bian, Z.~Lei, and S.~Z. Li.
\newblock Single-shot refinement neural network for object detection.
\newblock {\em arXiv preprint}, 2017.

\bibitem{dcnv2}
X.~Zhu, H.~Hu, S.~Lin, and J.~Dai.
\newblock Deformable convnets v2: More deformable, better results.
\newblock {\em arXiv preprint arXiv:1811.11168}, 2018.

\bibitem{edgebox}
C.~L. Zitnick and P.~Doll{\'a}r.
\newblock Edge boxes: Locating object proposals from edges.
\newblock In {\em European conference on computer vision}, pages 391--405.
  Springer, 2014.

\end{thebibliography}
}

\end{document}